\pdfoutput=1

\documentclass[11pt]{article}

\usepackage{acl}
\PassOptionsToPackage{table}{xcolor}
\usepackage{times}
\usepackage{latexsym}

\usepackage[T1]{fontenc}

\usepackage[utf8]{inputenc}

\usepackage{microtype}

\usepackage{inconsolata}

\usepackage{graphicx}
\usepackage{caption}
\usepackage{subcaption}
\usepackage{multicol}
\usepackage{makecell}
\usepackage{tikz}
\usepackage{arydshln}
\usepackage{array} 
\usepackage{multirow}
\usepackage{listings}
\usepackage{svg}
\usepackage{pifont}
\usepackage[breakable]{tcolorbox}
\usepackage{pgffor}
\usepackage{longtable}
\usepackage{algorithm}
\usepackage{algpseudocode}
\usepackage{amsmath}
\usepackage{bbm}
\usepackage{cleveref}
\usepackage{amssymb}
\usepackage{cleveref}
\usepackage{booktabs}
\usepackage{mathtools}

\usepackage{enumitem}
\usepackage{bm}
\usepackage{pgfplots}
\usepackage[table]{xcolor}

\NewDocumentCommand{\Blank}{ O{3} }{%
    \foreach \x in {0, 0.025, ..., #1}{%
        \rule{0.0125in}{0.5pt}%
        \hspace*{0.0125in}%
    }
}
\makeatletter
\def\addlegendimage{\pgfplots@addlegendimage}
\makeatother

\usetikzlibrary{patterns,patterns.meta}
\usepackage{filecontents}
\usepackage{pgfplots}
\usepgfplotslibrary{groupplots}
\usepackage{placeins}
\makeatletter
\algrenewcommand\ALG@beginalgorithmic{\footnotesize}
\makeatother

\usetikzlibrary{calc}
\usetikzlibrary{decorations.pathreplacing, positioning}

\usepackage{pgfplots}
\pgfplotsset{compat=1.12}

\title{Integrating Large Language Models with Graph-based  Reasoning for Conversational Question Answering}
\author{Parag Jain \qquad Mirella Lapata \\
Institute for Language, Cognition and Computation\\
School of Informatics, University of Edinburgh\\
\texttt{parag.jain@ed.ac.uk}~~~~\texttt{mlap@inf.ed.ac.uk}\\
}

\begin{document}
\interfootnotelinepenalty=10000

\definecolor{azure(colorwheel)}{rgb}{0.0, 0.5, 1.0}
\definecolor{coral}{rgb}{1.0, 0.5, 0.31}
\definecolor{darkcoral}{rgb}{0.8, 0.36, 0.27}
\definecolor{darkgoldenrod}{rgb}{0.72, 0.53, 0.04}
\definecolor{cadmiumgreen}{rgb}{0.0, 0.42, 0.24}
\definecolor{airforceblue}{rgb}{0.36, 0.54, 0.66}
\definecolor{cornflowerblue}{rgb}{0.39, 0.58, 0.93}
\definecolor{mediumseagreen}{RGB}{60, 179, 113}
\definecolor{seagreen}{rgb}{0.18, 0.55, 0.34}
\definecolor{lightsalmon}{RGB}{255, 160, 122}
\definecolor{plum}{RGB}{221, 160, 221}
\definecolor{darkseagreen}{rgb}{0.56, 0.74, 0.56}
\definecolor{darktan}{rgb}{0.57, 0.51, 0.32}
\definecolor{arylideyellow}{RGB}{255, 242 , 204}
\maketitle
\begin{abstract}
We focus on a conversational question answering task which combines the challenges of understanding questions in context and reasoning over  evidence gathered from heterogeneous sources like text, knowledge graphs, tables, and infoboxes.  Our method utilizes a graph structured representation to aggregate information about a question and its context (i.e., the conversation so far and evidence retrieved to find an answer), while also harnessing the reasoning and text generation capabilities of large language models (LLMs). Graph embeddings are directly injected into the LLM, bypassing the token embedding layers, and learned end-to-end by minimizing cross-entropy. Our model maintains a memory module to track and update past evidence, thus influencing the graph’s structure, as the
conversation evolves. Experimental results on the ConvMix benchmark \cite{Christmann_2022} show that graph embeddings enhance the LLM's ability to reason, while the memory module provides robustness against noise and retrieval errors.
\end{abstract}
\section{Introduction}
\label{sec:introduction}

Conversational question answering is an information seeking task where
users engage in interactive conversations with AI systems
\cite{choi-etal-2018-quac,reddy-etal-2019-coqa,Dalton:ea:2022}.
Unlike traditional question answering
applications~\cite{rajpurkar-etal-2016-squad}, conversational systems
are expected to track the context of a conversation, i.e.,~remember
previous questions and answers to provide relevant responses in an
ongoing dialogue. The majority of prior work has studied different
instantiations of conversational question answering, based on the
simplifying assumption that answers can be found in a \emph{single}
information source. Examples include querying knowledge graphs such as
Wikidata
\cite{perez-beltrachini-etal-2023-semantic,Christmann_2022,10.5555/3504035.3504122},
identifying answer spans in Wikipedia articles
\cite{reddy-etal-2019-coqa, choi-etal-2018-quac}, and searching for
answers in table cells \cite{iyyer-etal-2017-search}.

\begin{figure*}[t]
\centering
\includegraphics[width=1.05\textwidth]{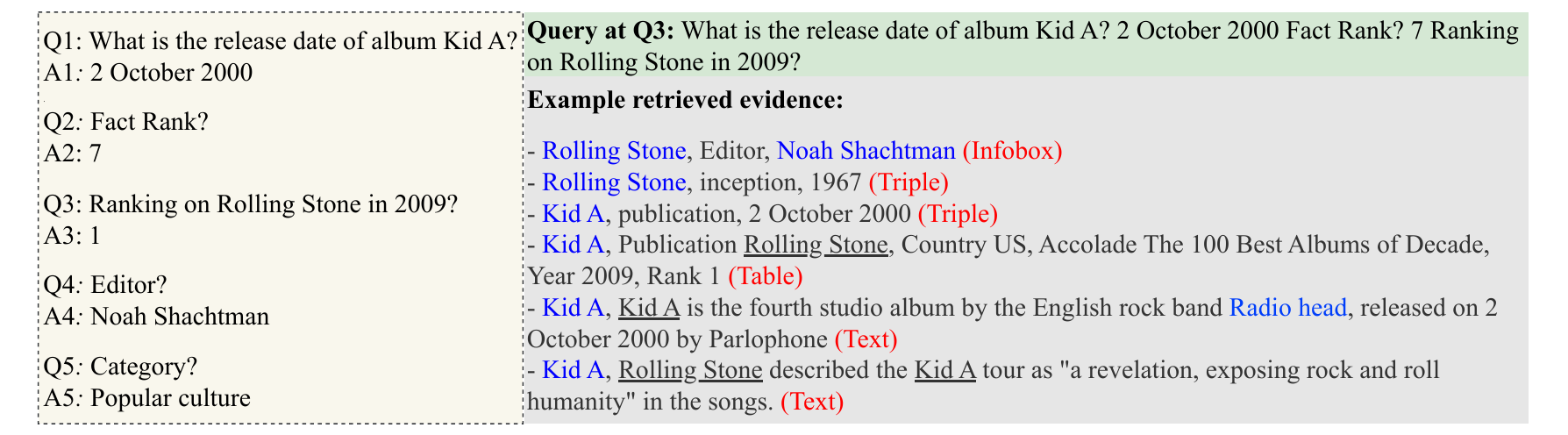}
\caption{Example interaction (left) from the ConvMix development set \cite{ConvMix}  and relevant evidence at query Q3 (right). Utterances Q1--Q3
  explore the topic of album \textsl{Kid A}. Q4 transitions to the
  topic of \textit{Rolling Stone} magazine. The evidence is retrieved from diverse  sources highlighted in \textcolor{red}{red}. Wikipedia text and tables are prepended with their respective article title. Known entities are shown in \textcolor{blue}{blue}. \underline{Underlined} entities are identified through string matching.}
\label{fig:example_interaction_source}
\end{figure*}

In this paper we focus on conversational question answering over
\emph{multiple} and \emph{heterogeneous} information
sources. Figure~\ref{fig:example_interaction_source} shows an example
interaction from ConvMix~\citep{ConvMix}, a recently curated dataset,
which combines the challenges of understanding questions in context,
and retrieving their answers from multiple sources. As can be seen,
answers are located in knowledge base triples (response to~Q1),
infoboxes (responses to Q4 and Q5), and tables (responses to Q2 and
Q3). It is also possible for an answer to be found in different
sources which may in turn disagree.  Moreover, the interaction in
Figure~\ref{fig:example_interaction_source} displays the hallmarks of
naturalistic dialogue. The second question (\textsl{Fact
  Rank?}) can only be interpreted by taking into account the topic of
the conversation (i.e.,~the album \textsl{Kid A}) mentioned in the
previous utterance.  Follow-on questions are short and may seem
ungrammatical taken out of context. As the conversation unfolds, the
topic shifts from the album \textsl{Kid A} to the \textsl{Rolling
  Stone} magazine; Q4 in Figure~\ref{fig:example_interaction_source}
has no dependencies on previous utterances and a hypothetical system
would have to recognize that a  new topic is being introduced.

\newcommand{\mybox}[2]{{\color{#1}\fbox{\normalcolor#2}}}

\begin{figure*}[t]
\centering
\includegraphics[width=0.95\linewidth]{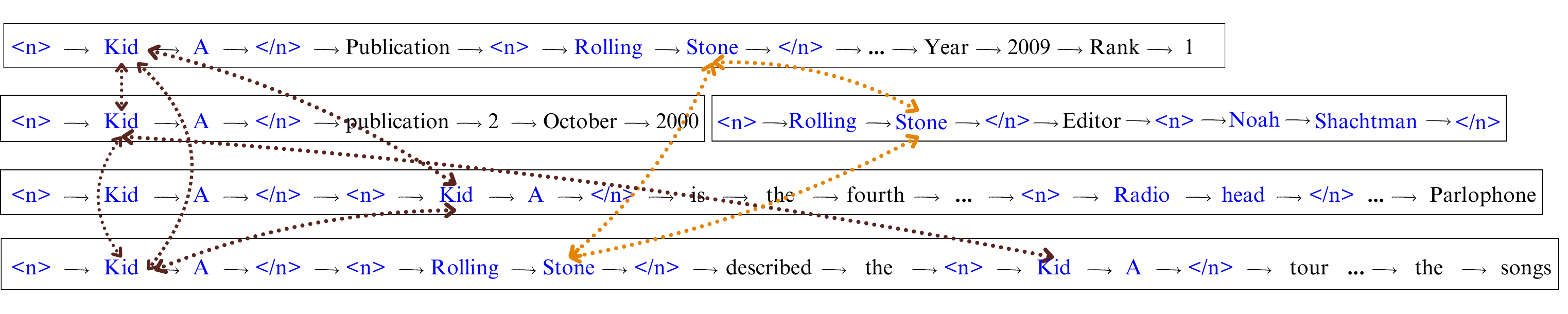}
\caption{Graph for retrieved evidence (subset) from
  Figure~\ref{fig:example_interaction_source}. Tokens within each
  \mybox{black}{instance} create local subgraphs in the form of a
  linear chain. Local subgraphs are connected through common
  entities (within <n> -- </n>) to build a global graph. Same
  color highlights connections between similar entities (some edges
  are omitted for   clarity).}
\label{fig:example_graph}
\end{figure*}

We propose a modeling approach to conversational question answering
which integrates large language models (LLMs) with graph-based
reasoning. The core idea is to represent information about a question
and its context --- such as the conversation so far and sources
retrieved to find an answer --- through a dynamically generated graph
and size  varies with each utterance.
Our method utilizes a graph structured representation
\cite{Gori:ea:2005,Scarselli:ea:2009} to aggregate information (and
resolve conflicts) from multiple sources, while also harnessing the
reasoning and text generation capabilties of LLMs.  Our graph network
is efficiently trained using gradients from the LLM. Graph
embeddings are directly injected into the LLM, bypassing the token
embedding layers, and learned end-to-end by minimizing cross-entropy
loss. To manage topic shifts and keep track of the conversation
flow, we introduce a \emph{memory} module that stores evidence used to
answer previous questions, thus allowing to re-use past information
for answering future questions. Our contributions are:
\begin{itemize}
  \setlength\itemsep{-0.1em}
  \vspace{-0.2cm}
\item A method  to aggregate evidence from
  multiple sources into a dynamic graph representation for 
  conversational question answering.
\item We efficiently integrate the evidence-based graph
  with LLMs for end-to-end training.
\item We keep track of past evidence in a memory module which is
  updated as the conversation evolves and  influences the graph
  structure and its representation. 
  \item Extensive experiments on the ConvMix dataset
    \cite{ConvMix}, demonstrate that graph structure enhances the
    LLM's ability to reason over multiple sources, while the
    memory module affords robustness to noise and retrieval errors.

  \end{itemize}


\section{Related Work}

\paragraph{Conversational Question Answering} 
Most previous work on conversational question answering operates over
a \emph{single} infromation source such as a knowledge graph, text
passage, or table~\citep{choi-etal-2018-quac,
  reddy-etal-2019-coqa,perez-beltrachini-etal-2023-semantic,iyyer-etal-2017-search}. Existing
models tend to be specialized, catering to isolated modalities
(e.g.,~text or tables), while a few approaches adopt graph-based
representations to organize the conversation and available
information~\citep{shen-etal-2019-multi,jain-lapata-2023-conversational,kacupaj-etal-2021-conversational,mueller-etal-2019-answering}.
A notable exception are ~\citet{EXPLAIGNN} who propose an end-to-end
model for \emph{multiple} information
sources. Specifically, their method constructs a heterogeneous graph
based on evidence retrieved from tables, infoboxes, text snippets, and
Wikidata triples.  This graph is iteratively pruned at inference time
to a smaller subgraph containing the answer (i.e., an entity node) to
the question.

Our work also integrates information from multiple sources into a
graph. However, we do not model question answering as a classification
task, but instead propose a generative model. We leverage  graph representations and the reasoning
capabilities of language models, without relying on specialized
inference procedures.

\paragraph{LLMs with Graphs}
A common approach to encoding graph structure for LLMs involves
describing the graph in natural language so that it resembles text
 \citep{ye2023natural,wang2024can}. There is no agreed consensus
on how to convert graphs to text, and most methods rely on
hand-crafted rules.  Previous efforts have shown it is
challenging for LLMs to reason over  graph
representations~\citep{fatemi2024talk,huang2024llms}, even when
explicit prompts are given that describe the structure of the graph in
natural language~\cite{huang2024llms}. Performance tends to be brittle
and task dependent \cite{wang2024can,fatemi2024talk}.

Our work proposes a parameter-efficient method for learning
\emph{task-specific} graph representations. It is closest
to~\citet{perozzi2024let}, who use graph embeddings as soft-prompts to
represent structured data for LLMs. In a similar vein,
\citet{chai2023graphllm} use prefix-tuning to integrate graph
embeddings with LLM attention layers. Their approach shows promising
results on small graphs with a few nodes ($\sim$20) and limited
variability. It also relies on the architecture of the LLM and may not
seamlessly integrate with other models, e.g.,~Mixture-of-Experts (MoE;
\citealt{shazeer2017outrageously, jacobs1991adaptive}).

\paragraph{Retrieval-augmented Generation}
Our work integrates LLMs with graph structural information 
 based on
evidence retrieved from the Wikidata knowledge
graph~\cite{10.1145/2629489}, Wikipedia text, tables, and
infoboxes. Although we do not focus on retrieval as such, it plays a
key role in identifying information for building the graph. Our
approach can thus be viewed as a variant of retrieval augmented
generation (RAG), since it conditions generation on freshly retrieved
evidence based on user
queries~\citep{10.5555/3648699.3648950,Khandelwal2020Generalization,guu2020retrieval}.


\section{Overview}
\label{sec:model-overview}

We assume a conversational question answering setting \cite{ConvMix}
that requires resoning over Wikipedia facts attested in multiple
sources such as text, tables, infoboxes, and the Wikidata knowledge
graph (KG). Given interaction~$\text{I}$, our task is to answer
question~$q_t$ at turn~$t$, taking into account retrieved
evidence~$r_t$ and previous turns $\text{I}[:t - 1]$ which consist of
questions and their answers~$\langle q_t, a_t \rangle$
(see Figure~\ref{fig:example_interaction_source}). To accomodate
information from the conversation so far, we concatenate question~$q_t$
at turn~$t$ with previous question-answer pairs, i.e.,~$Q_t = [q_1,
  a_1 \ldots q_{t-1}, a_{t-1}, q_t]$, and use this to retrieve evidence. 

\newcommand*\circled[1]{\tikz[baseline=(char.base)]{\node[shape=circle,draw,inner sep=0.8pt] (char) {#1};}}
\begin{figure*}[t]
    \centering
    \includegraphics[width=\textwidth]{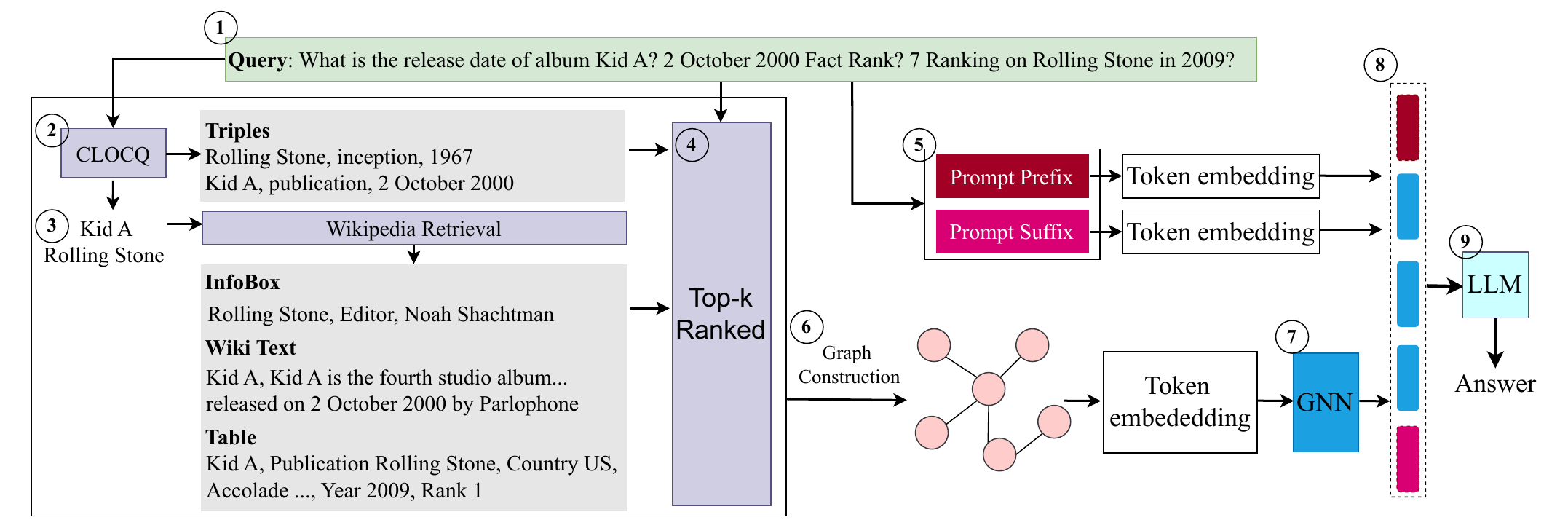}
    \caption{Sketch of proposed architecture. \protect\circled{1} shows  query Q3 from the interaction in Figure~\ref{fig:example_interaction_source}.
    \protect\circled{2} shows
       KG triples retrieved with CLOCQ and their  entities (\protect\circled{3}). Wikipedia articles for \protect\circled{3} are parsed to extract sentences, infoboxes and tables. In \protect\circled{4}, retrieved evidence is ranked based on the current query using BM25. \protect\circled{5} creates an instruction prompt based on the input
      query (see Appendix~\ref{app:sec:prompts} for the prompt template). In \protect\circled{6}, a graph is constructed based on top ranked instances. \protect\circled{7} depicts the learned graph neural network. Graph node embeddings are initialized using
      LLM token embeddings that are separate from the
      base model. \protect\circled{8} shows the final embeddings 
      which are passed to the LLM and are obtained by concatenating prompt (prefix, suffix) and graph  embeddings (shown in different colors). \protect\circled{9} is the LLM without the token
      embedding layer.} 
    \label{fig:model}
\end{figure*}

As depicted in Figure~\ref{fig:model}, we adopt a modular
approach. Given query~$Q_t$, we retrieve and rank relevant evidence
(Section~\ref{sec:retrieval}). We next organize retrieved information
into a graph (Section~\ref{sec:graph_construction}) and learn graph
embeddings using Graph Attention Networks (GAT;~\citealt{gat2018graph,
  brody2022how}). Finally, graph embeddings are injected in a LLM by
skipping the token embeddings layer
(Section~\ref{sec:graph_llm_integeration}). Unlike~\citet{EXPLAIGNN}
who \emph{extract} answers from retrieved evidence, we
\emph{generate} them.  Our model~$\mathcal{M}$ is thus formulated as:
\begin{equation}
    a_t = \mathcal{M} \left( \text{I}[:t - 1], q_t, r_t; \Theta \right)
\end{equation}
where~$q_t$ is the current question, $r_t$ is the graph representing
retrieved evidence, \mbox{$\text{I}[:t - 1]$} are previous turns, and $\Theta$ the parameters of our model which are fine-tuned  on task-specific data (Section~\ref{sec:training}).

\section{Model}
\subsection{Evidence Retrieval}
\label{sec:retrieval}

We adopt the retrieval pipeline outlined in~\citet{ConvMix}. As
mentioned earlier, information is obtained from Wikipedia pages and
the Wikidata KG using a query based on the current question
concatenated with previous question-answer pairs. Retrieval takes
place in two stages. Initially, evidence is retrieved from the
Wikidata KG, and then followed by retrieval from Wikipedia.

We extract Wikidata triples (see  \protect\circled{2} in Figure~\ref{fig:model}) using
CLOCQ~\citep{Christmann_2022}, a retrieval engine specifically
tailored to question answering over knowledge bases. It preprocesses
the knowledge graph in a memory efficient manner and returns the
top-$k$ triples based on query terms. Figure~\ref{fig:model}, shows a subset of relevant triples retrieved for~Q3 along with the KG entities~$E_{\mathcal{E}}$. 

We next obtain evidence pertaining to additional Wikipedia sources
by retrieving articles corresponding to the entities in~$E_{\mathcal{E}}$. These pages are subsequently processed to
extract text, tables, and infoboxes (see  \protect\circled{3} in  
Figure~\ref{fig:model}). Tables are linearized by individually
transforming each row into text and concatenating it with
corresponding column headers. 
 Infoboxes are linearized in a
similar fashion by concatenating key-value pairs with header
information (if available). 
KB triples are
linearized by a simple concatenation of individual elements. Wikipedia text is split into sentences, each of which serves as a separate piece of evidence.

The evidence collected at this stage can be extensive, potentially
comprising of several thousand instances, which
would in turn lead to a very large graph (see
Section~\ref{sec:graph_construction}).  To manage this, we employ
BM25~\citep{bm25} to rank the evidence against the query and retain
only the best scoring instances (see  \protect\circled{4} in Figure~\ref{fig:model}). Let~${E}_t$ denote the set of top-$k$
retrieved instances at turn~$t$.

\subsection{Evidence Memory}
\label{sec:memory}

By design, we retrieve new evidence at every turn~$t$, which may
suggest that every question introduces a new topic. However, a
well-known property of conversational dialogue is \emph{topic
  inertia}~\citep{chai-jin-2004-discourse}, i.e.,~users tend to
explore the same topic for a while before switching to a new topic
(see the interaction in Figure~\ref{fig:example_interaction_source}).
We propose to keep track of past topics through a memory module which
stores previously retrieved pieces of evidence to be re-utilized and
re-ranked against~$Q_t$. Specifically, at each turn~$t$ we define
evidence memory~$M_t$ as,
\begin{equation}
    M_t = \oplus\, \{E_j\, \mid \, j \in [1 \ldots t - 1] \}
\end{equation}
where~$\oplus$ denotes concatenation. We replace a proportion (e.g.,
one third) of low-ranked instances from~$E_t$ with the top-ranking
ones from~$M_t$. We employ the Sentence-BERT model
\cite{reimers-gurevych-2019-sentence} to re-rank the evidence stored
in $M_t$, using~$Q_t$ as a query.

\subsection{Graph Construction}
\label{sec:graph_construction}
Retrieved information is organized into a graph  (see~\protect\circled{6},  Figure~\ref{fig:model}) by first converting
individual pieces of evidence into a linear chain. \emph{Local} 
subgraphs are then  merged into a \emph{global} 
graph by linking common entities between them. Figure~\ref{fig:example_graph} 
shows example graphs with \emph{local} and \emph{global} connections.

To construct a \emph{local} graph, evidence from different sources is linearized (as discussed in Section~\ref{sec:retrieval}) and tokenized using a
base LLM tokenizer. Tokens within each instance are treated as graph nodes connected in
a linear chain. In other words, evidence~$w$ with tokens~$w_1 \hdots
w_{|w|}$ is represented by local sub-graph $w_1 \rightarrow w_2
\rightarrow \hdots \rightarrow w_{|w|}$.  

Connecting different pieces of evidence together is critical for
enabling more global reasoning. We create a \emph{global} graph by linking similar entities across \emph{local} subgraphs. In this context,  entities are KG items  but also text spans in Wikipedia text, infoboxes, and tables gathered during retrieval. We identify entity spans by performing string matching against KG entities. In Figure~\ref{fig:example_graph}, such entities are encircled by \texttt{<n>} \textit{node} \texttt{</n>} tags. Finally, entity spans referring to same entity are  linked, thus creating a more globally connected graph. 
\subsection{Graph Encoder}
\label{sec:graph:sencoder}

Our model generates an answer at each turn~$t$ given query $Q_t$ and
graph~\mbox{$\mathcal{G_\text{t}}$} representing relevant evidence
(see Figure~\ref{fig:model}). More formally,
\mbox{$\mathcal{G_\text{t}} = (\mathcal{V}, \mathcal{E})$} is a
directed graph with nodes~$\mathcal{V} = \{v_1, v_2, \dots , v_n\}$
and edges~$\mathcal{E} \subseteq \mathcal{V} \times \mathcal{V}$.

We do not learn graph node embeddings from scratch. Instead, we
initialize them using token embeddings from a large language
model (see~\protect\circled{7}, Figure~\ref{fig:model}). This step is crucial for achieving feature alignment between
the evidence graph and the downstream LLM. Generally, integrating LLMs
with information from a different modality necessitates aligning
features between them. For example, vision-language models like
BLIP-2~\citep{li2023blip} and LLaVA~\citep{NEURIPS2023_6dcf277e}
perform feature alignment by heavily pretraining a network whose goal
is to act as a bridge between a frozen image encoder and a frozen
LLM. This approach requires large amounts of pretraining data (as well
as computational resources) which are not readily available for our
task. We found that simply initializing graph node embeddings with
token embeddings from a base LLM is effective and crucial for
achieving good performance.

Let $\{x_{i} \mid i \in [1, n]\}$ denote the set of initial node
embeddings. We learn
graph structure representations with the Graph Attention Network (GAT;
\citealt{gat2018graph, brody2022how}), a neural network architecture
designed for handling graph-structured data.  It is
computationally efficient, it requires less memory and storage
compared to other deep learning models, and is applicable to inductive
problems. GAT uses the attention mechanism to weigh the importance of
neighboring nodes when aggregating information in a graph.  Attention
between two nodes is calculated as:
\begin{equation}\label{attncoeff}
    \alpha_{ij} = \frac{\exp\bigl( \psi \bigl( x_i, x_j \bigr) \bigr)}{\sum_{k\in\mathcal{N}_i}\exp\bigl(\psi\bigl(x_i, x_k\bigr)\bigr)}
\end{equation}
where $\mathcal{N}_i = \{ v_j \in \mathcal{V} \mid \bigl( j,i \bigr)
\in \mathcal{E} \} $ are the neighbors of node~$v_i$,
and~$\alpha_{ij}$ is the attention score between node embeddings~$x_i$
and~$x_j$.  Following~\citet{brody2022how}, we compute the scoring
function~$\psi$ as:
\begin{multline}
  \hspace*{-.2cm}   \psi \bigl( x_i , x_j \bigr) = a^T\operatorname{LeakyReLU}\bigl( W \cdot [x_i \oplus x_j]\bigr)
     \end{multline}
where~$\cdot^T$ represents transposition and~$\oplus$ is the
concatenation operation. Attention coefficients corresponding to each
node~$i$ are then used to compute a linear combination of the features
corresponding to neighboring nodes as:
\begin{equation}\label{eqnatt}
	x_i = \sigma\left(\sum_{j\in\mathcal{N}_i} \alpha_{ij} {W}x_j\right)
\end{equation}

\subsection{Integration with LLMs}
\label{sec:graph_llm_integeration}

The LLM takes as input a composite embdedding consisting of the graph
embeddings discussed above, and embeddings corresponing to a prompt
prefix~$\mathsf{P}_{\text{prefix}}$, and a prompt suffix~$\mathsf{P}_{\text{suffix}}$
(see \protect\circled{5} in Figure~\ref{fig:model}). $\mathsf{P}_{\text{prefix}}$ is an initial instruction prompt and $\mathsf{P}_{\text{suffix}}$ represents the conversational query at turn~$t$  to be answered. See Appendix~\ref{app:sec:prompts}(Figure~\ref{fig:graph_prompt})  for an example prompt. More formally, LLM input
embeddings are obtained as:
\begin{equation}
    \label{eq:concat_emb}
    \mathsf{H} = \mathsf{H}_{\text{prefix}} \oplus \mathsf{H}_g \oplus \mathsf{H}_{\text{\text{suffix}}}
\end{equation}
where $\mathsf{H}_g$ is the list of embeddings of all graph
nodes and $\mathsf{H}_{\text{prefix}}$ is the text embedding of
$\mathsf{P}_{\text{prefix}}$:
\begin{equation}
\label{eq:prefix_token_encoding}
  \mathsf{H}_{\text{prefix}} =  \operatorname{Embed}(\operatorname{Tok}(\mathsf{P}_{\text{prefix}}))
\end{equation}
where~ $\operatorname{Tok}$ and $\operatorname{Embed}$ are the base
LLM tokenizer and embedding layer, respectively. ~$\mathsf{P}_{\text{suffix}}$ is encoded in a similar manner using Equation~\eqref{eq:prefix_token_encoding} to obtain $\mathsf{H}_{\text{suffix}}$. We use the embeddings obtained with
Equation~(\ref{eq:concat_emb}) as the initial token embeddings for the
pretrained LLM.

\subsection{Training}
\label{sec:training} 

Our model is trained end-to-end by optimizing cross-entropy loss.  For
all variants (with and without graph structure), the loss is
calculated on completion tokens only, i.e., prompt tokens do not
observe any loss. This is similar to setting the prompt loss weight to
0~\citep{wang-etal-2023-self-instruct}.

Given  training instance $\langle \text{I}[:t-1], q_t, r_t;
\Theta \rangle$, and  sequence of gold output tokens
$\langle a^1_t, a^2_t, \dots ,a^{|a_t|}_t \rangle$, we minimize 
token-level cross-entropy as:
\begin{equation}
    \mathcal{L} \left( \hat{a}^i_t \right) = - \operatorname{log}p \left( a^i_t \mid \text{I}[:t - 1], q_t, r_t; \Theta \right)
\end{equation}
where~$\hat{a}^i_t$ denotes the predicted output token at decoder step~$i$.
We use a mixed approach for training the whole network. Our graph
network is trained from scratch, however, the base LLM is updated
using LoRA~\citep{hu2022lora} in a parameter efficient manner. We perform inference based on the conversation context (i.e., $\text{I}[:t-1]$) and current query~$q_t$.


\section{Experimental Setup}
\label{sec:experimental-setup}

We use \mbox{Mistral-7B-Instruct-v0.2} \cite{jiang2023mistral} as our base model, given its good performance across complex reasoning tasks, and wider context window of~32K tokens. Recall that we retrieve and encode a large
number of instances as evidence for a question. 
Our implementation predominantly relies on
PyTorch~\citep{paszke2019pytorch}. We adapt the Mistral implementation
available at the HuggingFace Transformers
library~\citep{wolf-etal-2020-transformers}. For developing the graph
neural network, we utilize PyTorch Geometric (PyG;
\citealt{Fey/Lenssen/2019}). We use Hugging Face's TRL
(Transformer Reinforcement Learning) library~\citep{vonwerra2022trl} for fine-tuning model without graph.
Additional training parameters and prompts can be found in Appendices ~\ref{app:sec:hyper} and ~\ref{app:sec:prompts}, respectively.

\subsection{Dataset}
\label{sec:dataset}

\begin{table}[t]
  \small
  \centering
\begin{tabular}{@{}l@{~}c@{}}
\toprule
\multicolumn{2}{c}{ConvMix-5T}      \\\hline
  Entities covered   & 5,418                                    \\
Long-tail entities & 2,511                                 \\
conversations      & 2,800                                    \\
Number of turns    & 5                                       \\
Split ratio             & 60:20:20 \\
\hline
\multicolumn{2}{c}{ConvMix-10T test set}      \\
\hline
Conversations      & 200                                     \\
Number of turns    & 10
\\ \midrule
\multicolumn{2}{@{}l@{}}{Domains: Books, Movies, Music,  TV series, Soccer} \\
\multicolumn{2}{@{}l@{}}{Answer Source: Text, Tables, Infobox Wikidata}\\ 
\bottomrule
\end{tabular}
\caption{ConvMix dataset statistics. Long tail entities are those
  attested in less than~50 KG facts.}
\label{tab:dataset}
\end{table}

We evaluate our work on ConvMix~\citep{ConvMix}, a conversational
question answering dataset that requires reasoning over heterogeneous
sources, specifically Wikipedia text, infoboxes, tables, and the
Wikidata KG. Aside from reasoning, the conversational nature of ConvMix
requires handling discourse phenomena, such as coreference, ellipsis,
and topic-shift~\citep{sun2007discourse, 10.1162/tacl_a_00422}.
Table~\ref{tab:dataset} summarizes various dataset statistics. As can
be seen (first block), the main dataset (CovMix-5T) contains~2,800
conversations, each with five turns (i.e.,~question-answer pairs),
split into training, development, and test set. 
In addition, ConvMix-10T is  a \emph{separate} test set used to measure
generalization on longer interactions. It contains 200 conversations,
each~10 turns long (see last block in Table~\ref{tab:dataset}).  We
follow the splits provided in~\citet{ConvMix} and report results on both
test sets combined.

\subsection{Evaluation Metrics}

Our model generates answers which may be valid but not identical to
the gold standard (e.g.,~\textsl{United States}, \textsl{United States
  of America}, and \textsl{USA} are all paraphrases of the same
concept).  When there is no exact match, we follow previous
work~\cite{ConvMix} and try to normalize the answer to its canonical
form. We use the Levenshtein distance~\citep{Levenshtein1965BinaryCC}
to measure the similarity of the generated answer with entities in our
retrieved evidence set. The entity with the smallest distance is used
as the answer in such cases. We report H@1 (i.e.,~precision at 1) and
H@5 (i.e.,~whether an answer match is found within the top 5 matching
entities).

\section{Results}
\label{sec:results}

Our experiments were designed to assess whether graph structure
enhances LLM performance for our conversational question-answering
task.  Our results are summarized in Table~\ref{tab:mistral_results}.

We evaluate our approach against Mistral-7B variants without graph
structure. Specifically, we compare against (a)~Mistral-7B zero-shot
prompted with top-$k$ retrieved instances and the conversational
history, i.e.,~the current query concatenated with previous QA pairs
(see Appendix~\ref{app:sec:prompts} for the prompt); and
(b)~Mistral-7B fine-tuned on the ConvMix training set using LoRA
(Mistral-7B + FT) and top-$k$ retrieved instances. We present three
variants of our model, fine-tined with graph embeddings (\mbox{Mistral-7B} + Graph) and
additionally with a memory management component (+Memory, +Rand
Memory).

We also compare with several state-of-the-art systems built on top of
T5~\cite{JMLR:v21:20-074}. T5-FiD \citep{ConvMix} is a
fusion-in-decoder model which acts as a ``generative reader'' and is
trained on (top-$k$) retrieved instances and gold
answers. Specifically, query-evidence pairs are encoded independently,
and passed on to the decoder to generate an answer. We also report
results with a T5-based model (T5-FiD + Question rewriting) which
rewrites the question based on the conversational history context
\cite{raposo2022question, elgohary-etal-2019-unpack} and a related
approach (\mbox{T5-FiD} + Question resolution) which performs query resolution,
i.e.,~by appending relevant terms from previous question-answer pairs
to the current question~\citep{voskarides2020query}.\footnote{All FiD models are based on T5-base~\citep{ConvMix}.}

Finally, although not directly comparable, we report the performance
of EXPLAIGNN~\citep{EXPLAIGNN} and Convinse T5-FiD~\cite{ConvMix}.
EXPLAIGNN is a classification model that identifies
entity nodes in a graph as answer predictions. It learns a
task specific structured representation optimized for better
retrieval and query understanding. The learned representation is used to train a
classification model based on graph neural networks tying both of them together. Convinse T5-FiD is similar in that it also learns a task-specific structured representation for retrieval and query understanding, without, however, creating a graph.

All models in Table~\ref{tab:mistral_results} use the same retrieval
engine (i.e., CLOCQ; \citealt{Christmann_2022}) which allows us to
focus on architectural differences and compare models on equal
footing.

\begin{table}[]
\resizebox{\linewidth}{!}{%
\begin{tabular}{lcc}
\toprule
\multicolumn{1}{c}{Models}    &                              H@1   & H@5                   \\ \midrule

Mistral-7B zero-shot   & 0.292 & 0.346                 \\ 
Mistral-7B + FT                          & 0.350  & 0.400                   \\
Mistral-7B + Graph                       & 0.425 & 0.459                 \\
Mistral-7B + Graph + Memory       & \textbf{0.445} & \textbf{0.512}                \\
Mistral-7B + Graph + Rand Memory   & 0.425 & 0.461                 \\ \midrule

T5-FiD           & 0.300   & 0.350 \\
T5-FiD + Question resolution &	0.282 &	0.297 \\
T5-FiD + Question rewriting 	& 0.271	& 0.285 \\ \midrule

Convinse T5-FiD &	0.342	& 0.386 \\
EXPLAIGNN &	0.406	& 0.561 \\ \bottomrule
\end{tabular}
}
\caption{Model performance on the ConvMix dataset (results are averaged
  for ConvMix-5T and convMix-10T test sets). H@1 represents precision
  at 1 and H@5 represents a match at 5. A fine-tuned Mistral-7B with graph embeddings and a memory module performs best.}
\label{tab:mistral_results}
\end{table}

\paragraph{Integrating LLMs with graph-based reasoning boosts
  conversational QA performance.}\label{sec:results_structure} As
shown in Table~\ref{tab:mistral_results}, Mistral-7B + Graph is
superior to a plain fine-tuned version of Mistral-7B (+ FT) by a large
margin. This suggests that organizing and representing retrieved
evidence as a graph improves reasoning compared to processing pieces
of evidence independently. Perhaps unsurprisingly, fine-tuning
generally improves Mistral's performance on the conversational QA task
over a zero-shot model. This is due to an improved understanding of
task requirements, like regular shift in focus and answer format. For example, the model learns to avoid verbosity in answers and respond using dataset-specific conventions such as  spelling  out the month in dates (e.g.,~\textsl{2 October 2002} instead of \textsl{2/10/2002}). The performance of the \mbox{T5-FiD}
systems is comparable to zero-shot Mistral-7B. In general, we observe
that performance improvements are not simply due to increased model
size. Rather, it is important to model the conversational nature of
the task and interpret the retrieved information more globally.

\paragraph{Adding a memory module improves QA precision.}\label{sec:results_memory}

Table~\ref{tab:mistral_results} shows that results further improve
when a memory module is added to our  model (+Graph +Memory). Recall that previously retrieved instances are kept in memory
and re-reranked against the current query. To further assess the
usefulness of re-ranking, we conducted a controlled experiment where
evidence was selected randomly from the memory. We observe that random
selection (+Rand Memory) amounts to not having a memory component at
all.

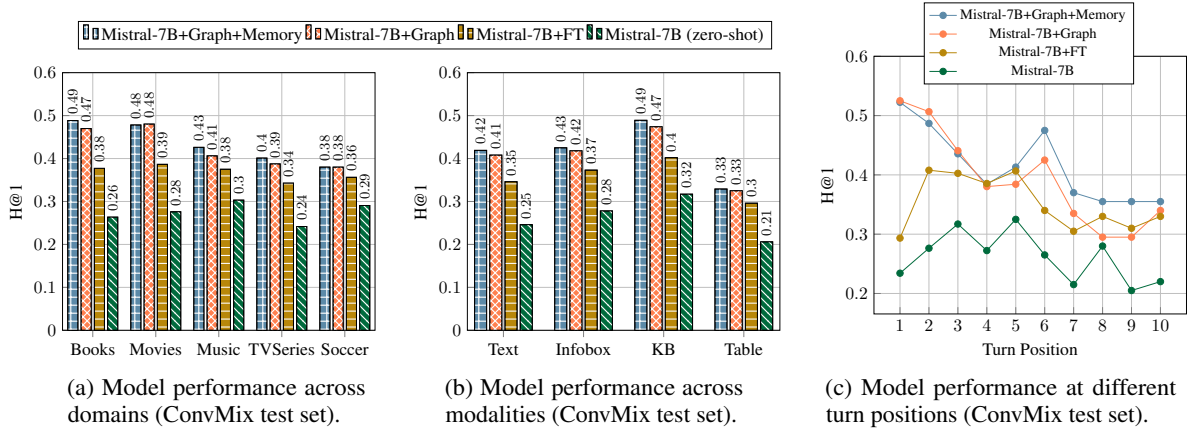
\begin{figure*}[t]
  \begin{subfigure}[t]{0.35\textwidth}
      \centering\captionsetup{width=.7\linewidth}
      \begin{tikzpicture}[scale=0.6]
    \begin{axis}[
        ybar,
        xtick pos=left,
        ytick pos=left,
        bar width=0.16,
        enlarge x limits=0.12,
         legend style={at={(2.27,1.2)},  legend columns=-1},
        ylabel={H@1},
        xticklabels={Books, Movies, Music, TVSeries, Soccer},
        xtick={0,1,2,3,4},
        ytick
        distance=0.1, ymax=0.6, ymin=0, grid=both,
        nodes near coords,
        every node near coord/.append style={rotate=90, anchor=west, text=black, font=\small},
        ]
        \addplot[black, fill=airforceblue,
           postaction={pattern=grid, pattern color=white}] table [x  expr=\coordindex, y index=1, col sep=comma] {domaindata.csv};
        
        \addplot[black, fill=coral,
           postaction={pattern=crosshatch, pattern color=white}] table [x  expr=\coordindex, y index=2, col sep=comma] {domaindata.csv};
        
        \addplot[black, fill=darkgoldenrod,
           postaction={pattern=horizontal lines, pattern color=white}] table [x  expr=\coordindex, y index=3, col sep=comma] {domaindata.csv};
        
        \addplot[black, fill=cadmiumgreen,
           postaction={pattern=north west lines, pattern color=white}] table [x  expr=\coordindex, y index=4, col sep=comma] {domaindata.csv};
        \legend{Mistral-7B+Graph+Memory~, Mistral-7B+Graph~, Mistral-7B+FT~, Mistral-7B (zero-shot)}
    \end{axis}
\end{tikzpicture}
\caption{Model performance across domains (ConvMix test set).}
\label{fig:domain_performance}
\end{subfigure}
\hspace*{-.8cm}
\begin{subfigure}[t]{0.35\textwidth}
  \centering\captionsetup{width=.7\linewidth}
  \begin{tikzpicture}[scale=0.6]
    \begin{axis}[
        ybar,
        xtick pos=left,
        ytick pos=left,
        bar width=0.15,
        enlarge x limits=0.15,
        ylabel={H@1},
        xticklabels={Text, Infobox, KB, Table},
        xtick={0,1,2,3},
        ytick
        distance=0.1, ymax=0.6, ymin=0, grid=both,
        nodes near coords,
        every node near coord/.append style={rotate=90, anchor=west, text=black, font=\small},
        ]
        
        \addplot[black, fill=airforceblue,
           postaction={pattern=grid, pattern color=white}] table [x  expr=\coordindex, y index=1, col sep=comma] {sourcedata.csv};
        \addplot[black, fill=coral,
           postaction={pattern=crosshatch, pattern color=white}] table [x  expr=\coordindex, y index=2, col sep=comma] {sourcedata.csv};
        \addplot[black, fill=darkgoldenrod,
           postaction={pattern=horizontal lines, pattern color=white}] table [x  expr=\coordindex, y index=3, col sep=comma] {sourcedata.csv};
        \addplot[black, fill=cadmiumgreen,
           postaction={pattern=north west lines, pattern color=white}]
        table [x  expr=\coordindex, y index=4, col sep=comma]
        {sourcedata.csv};
    \end{axis}
\end{tikzpicture}
    \caption{Model performance across  modalities (ConvMix test set).}
    \label{fig:source_performance}
\end{subfigure}
\hspace*{-.5cm}
\begin{subfigure}[t]{0.36\textwidth}
  \centering\captionsetup{width=.8\linewidth}
    \begin{tikzpicture}[scale=0.6]
    \begin{axis}[xlabel={Turn Position},ylabel={H@1},tick pos=left, legend
        style={at={(0.16,1.05)},anchor=west},xtick={1,2,3,4,5,6,7,8,9,10},ytick
        distance=0.1, ymax=0.6, grid=both, align=right] 
    \addplot[airforceblue,mark=*, mark size=2pt] table[x index=0,y index=1,col sep=comma] {len_data.dat};
    \addlegendentry{\small Mistral-7B+Graph+Memory}

    \addplot[coral,mark=*, mark size=2pt] table[x index=0,y index=2,col sep=comma] {len_data.dat};
    \addlegendentry{\small Mistral-7B+Graph} 

     \addplot[darkgoldenrod,mark=*, mark size=2pt] table[x index=0,y index=3,col sep=comma] {len_data.dat};
    \addlegendentry{\small Mistral-7B+FT}
    
    \addplot[cadmiumgreen,mark=*, mark size=2pt] table[x index=0,y index=4,col sep=comma] {len_data.dat};
    \addlegendentry{\small Mistral-7B}
    
    \end{axis}
    \end{tikzpicture}
    \caption{Model performance at different turn positions (ConvMix
      test set).}
    \label{fig:len_chart}
\end{subfigure}
\caption{Analysis experiments for different model variants based on Mistral-7B prompted in a zero-shot setting, fine-tuned on ConMix without graph embeddings (+FT), with graph embeddings (+Graph), and with a memory module (+Graph +Memory). Performance degrades with numbers, tables, and later conversation turns.}
\end{figure*}

\paragraph{It is challenging to provide accurate answers to questions
  that require numerical responses.}
Figure~\ref{fig:domain_performance} shows model performance broken down by question domain. Overall, we observe similar trends
across domains, with \texttt{TV Series} and \texttt{Soccer} being most
challenging. Performance for these domains decreases by $\sim$10
percentage points, e.g.,~in comparison to \texttt{Books}.  To uncover
the reason for this gap, we further investigate whether there is an
effect of answer type. We automatically annotate\footnote{We use
\texttt{regex} and \texttt{python-dateutil} to automatically
categorize the answers.}  the ConvMix development set with the
following answer categories: strings, dates, and numbers. The results
in Table~\ref{tab:numeric_results} (top) show average H@1 stratified
by different answer types.

We observe that questions with numeric
answers are harder compared to other categories. There are several 
reasons for this, including  variability in numerical reasoning performance due to the choice of numeric data tokenization by the base model~\citep{singh2024tokenization,sun-etal-2023-tokenization}. As well as the effect of pre-training data on the  output predictions and their  probability~\citep{mccoy2023embers}.
Table~\ref{tab:numeric_results}b (bottom) reveals that the proportion
of instances with numeric answers is highest for the \texttt{TV
  Series} and \texttt{Soccer} domains, thus explaining why 
performance drops for these domains.

\begin{table}[t]
  \small
  \centering
\raisebox{.7em}[0pt]{(a)}\hspace*{1em} \begin{tabular}{@{}c|ccc@{}} \toprule
  Answer Type & Date & String & Number  \\ 
H@1  & 0.50 & 0.45 & 0.14 \\ \bottomrule
\multicolumn{4}{@{}c@{}}{} \\
\end{tabular}

(b)\hspace*{1em} \begin{tabular}{@{}c@{~}|c@{~}c@{~~}c@{~~}c@{~~}c@{}} \toprule
Domain & Books & Movies & Music & TV series & Soccer \\ 
\% Number &  3.9 & 2.1 & 5.0& 10.0 & 7.9 \\\bottomrule
\end{tabular}
\caption{\label{tab:numeric_results} Model performance (Mistral-7B +
  Graph + Memory) across answer types (top) and proportion of numeric
  answers per domain (ConvMix dev set).}
\end{table}

\paragraph{It is challenging to extract accurate information from tables.}
Figure~\ref{fig:source_performance}, shows how  performance
varies depending on the source of the answer. Across models, we observe that
performance deteriorates when the answers are located in tables.  On
the contrary, performance is generally better when answers are found
in the knowledge graph. We believe this performance gap is due to how
tabular information is linearized. In contrast to the knowledge graph
from which facts can be easily extracted, Wikipedia tables often have
complex hierarchical structure \cite{parikh-etal-2020-totto} making it
challenging to achieve clean and robust
linearization~\cite{alonso2023pixt3}.

\paragraph{It is more difficult to answer questions 
  occurring later in the conversation.} In Figure~\ref{fig:len_chart}
we examine how performance varies with conversation length.
Ideally, a model should be able to answer questions irrespective of
where these occur (e.g.,~beginning or end). As mentioned in
Section~\ref{sec:dataset}, ConvMix contains conversations with a
maximum length of 10~turns.  The results in Figure~\ref{fig:len_chart}
show a general decrease in performance as the dialogue progresses.
Initial questions tend to be more complex while follow-on questions
often extend or elaborate upon the initial
topic~\citep{chai-jin-2004-discourse, 10.1162/tacl_a_00422}. Our
results  show that graph enhanced models
generally outperform LLM variants which do not organize the retrieved
information in any way. Furthermore, we observe that having a memory
(of previously retrieved instances) is particularly helpful in longer
interactions. Keeping track of past evidence helps ameliorate
retrieval errors which might erroneously steer the model towards new
topics. Aside from contextual factors, the quality of retrieval
largely influences model precision, as approximately half of the
answers cannot be found even at the beginning of the dialogue (see
turn 1 in Figure~\ref{fig:len_chart}).


\section{Conclusion}
In this paper we propose a method to aggregate evidence from multiple
sources into a dynamic graph representation for conversational
question answering. We demonstrate how this graph can be efficiently
integrated with large language models (LLMs) for end-to-end training,
enhancing the model's ability to handle evolving conversational
contexts. Our approach maintains a memory module to track and update past
evidence, thus influencing the graph's structure and representation,
as the conversation evolves.  Experiments on the ConvMix
dataset show that the graph enhances the LLM's
ability to reason over multiple modalities, while the memory module
provides robustness against noise and retrieval errors. In
the future, we would like to improve information retrieval for our task, through using pretrained embeddings for better entity linking. We could also adopt a structured memory module for more complex reasoning. 


\section{Limitations}
Our experiments are limited to one dataset (i.e.,~ConvMix) and one language, namely English. It would be interesting to see if our findings generalize to other datasets which are conversational in nature but do not target our specific question answering task. For example, $\mathbb{SPICE}$~\citep{perez-beltrachini-etal-2023-semantic} is a
recently released conversational semantic parsing dataset where utterances are translated into  executable semantic parses (in this
case \textsc{Sparql} queries). It would also be interesting to examine how our model handles languages other than English, however, we are not aware of any multi-lingual or cross-lingual datasets for conversational question answering. 

In this work, we do not study the effect of various prompting techniques  on our task. In experiments, we found Mistral-7B's performance superior to Llama2-7B \cite{touvron2023llama}, however, we did not perform an in-depth study on prompts and models.  Measuring the effect of these factors on our task and model performance is non-trivial and a topic for future work.

\bibliography{anthology,custom}

\begin{thebibliography}{56}
\providecommand{\natexlab}[1]{#1}

\bibitem[{Alonso et~al.(2023)Alonso, Agirre, and Lapata}]{alonso2023pixt3}
I{\~n}igo Alonso, Eneko Agirre, and Mirella Lapata. 2023.
\newblock Pixt3: Pixel-based table to text generation.
\newblock \emph{arXiv preprint arXiv:2311.09808}.

\bibitem[{Brody et~al.(2022)Brody, Alon, and Yahav}]{brody2022how}
Shaked Brody, Uri Alon, and Eran Yahav. 2022.
\newblock \href {https://openreview.net/forum?id=F72ximsx7C1} {How attentive are graph attention networks?}
\newblock In \emph{The Tenth International Conference on Learning Representations, {ICLR} 2022, Virtual Event, April 25-29, 2022}. OpenReview.net.

\bibitem[{Chai and Jin(2004)}]{chai-jin-2004-discourse}
Joyce~Y. Chai and Rong Jin. 2004.
\newblock \href {https://aclanthology.org/W04-2504} {Discourse structure for context question answering}.
\newblock In \emph{Proceedings of the Workshop on Pragmatics of Question Answering at {HLT}-{NAACL} 2004}, pages 23--30, Boston, Massachusetts, USA. Association for Computational Linguistics.

\bibitem[{Chai et~al.(2023)Chai, Zhang, Wu, Han, Hu, Huang, and Yang}]{chai2023graphllm}
Ziwei Chai, Tianjie Zhang, Liang Wu, Kaiqiao Han, Xiaohai Hu, Xuanwen Huang, and Yang Yang. 2023.
\newblock Graphllm: Boosting graph reasoning ability of large language model.
\newblock \emph{arXiv preprint arXiv:2310.05845}.

\bibitem[{Choi et~al.(2018)Choi, He, Iyyer, Yatskar, Yih, Choi, Liang, and Zettlemoyer}]{choi-etal-2018-quac}
Eunsol Choi, He~He, Mohit Iyyer, Mark Yatskar, Wen-tau Yih, Yejin Choi, Percy Liang, and Luke Zettlemoyer. 2018.
\newblock \href {https://doi.org/10.18653/v1/D18-1241} {{Q}u{AC}: Question answering in context}.
\newblock In \emph{Proceedings of the 2018 Conference on Empirical Methods in Natural Language Processing}, pages 2174--2184, Brussels, Belgium. Association for Computational Linguistics.

\bibitem[{Christmann et~al.(2022{\natexlab{a}})Christmann, Saha~Roy, and Weikum}]{Christmann_2022}
Philipp Christmann, Rishiraj Saha~Roy, and Gerhard Weikum. 2022{\natexlab{a}}.
\newblock \href {https://doi.org/10.1145/3488560.3498488} {Beyond ned: Fast and effective search space reduction for complex question answering over knowledge bases}.
\newblock In \emph{Proceedings of the Fifteenth ACM International Conference on Web Search and Data Mining}, WSDM ’22. ACM.

\bibitem[{Christmann et~al.(2022{\natexlab{b}})Christmann, Saha~Roy, and Weikum}]{ConvMix}
Philipp Christmann, Rishiraj Saha~Roy, and Gerhard Weikum. 2022{\natexlab{b}}.
\newblock \href {https://doi.org/10.1145/3477495.3531815} {Conversational question answering on heterogeneous sources}.
\newblock In \emph{Proceedings of the 45th International ACM SIGIR Conference on Research and Development in Information Retrieval}, SIGIR '22, page 144–154, New York, NY, USA. Association for Computing Machinery.

\bibitem[{Christmann et~al.(2023)Christmann, Saha~Roy, and Weikum}]{EXPLAIGNN}
Philipp Christmann, Rishiraj Saha~Roy, and Gerhard Weikum. 2023.
\newblock \href {https://doi.org/10.1145/3539618.3591682} {Explainable conversational question answering over heterogeneous sources via iterative graph neural networks}.
\newblock In \emph{Proceedings of the 46th International ACM SIGIR Conference on Research and Development in Information Retrieval}, SIGIR '23, page 643–653, New York, NY, USA. Association for Computing Machinery.

\bibitem[{Dalton et~al.(2022)Dalton, Fischer, Owoicho, Radlinski, Rossetto, Trippas, and Zamani}]{Dalton:ea:2022}
Jeffrey Dalton, Sophie Fischer, Paul Owoicho, Filip Radlinski, Federico Rossetto, Johanne~R. Trippas, and Hamed Zamani. 2022.
\newblock \href {https://doi.org/10.1145/3477495.3532678} {Conversational information seeking: Theory and application}.
\newblock In \emph{Proceedings of the 45th International ACM SIGIR Conference on Research and Development in Information Retrieval}, SIGIR '22, page 3455–3458, New York, NY, USA. Association for Computing Machinery.

\bibitem[{Elgohary et~al.(2019)Elgohary, Peskov, and Boyd-Graber}]{elgohary-etal-2019-unpack}
Ahmed Elgohary, Denis Peskov, and Jordan Boyd-Graber. 2019.
\newblock \href {https://doi.org/10.18653/v1/D19-1605} {Can you unpack that? learning to rewrite questions-in-context}.
\newblock In \emph{Proceedings of the 2019 Conference on Empirical Methods in Natural Language Processing and the 9th International Joint Conference on Natural Language Processing (EMNLP-IJCNLP)}, pages 5918--5924, Hong Kong, China. Association for Computational Linguistics.

\bibitem[{Fatemi et~al.(2024)Fatemi, Halcrow, and Perozzi}]{fatemi2024talk}
Bahare Fatemi, Jonathan Halcrow, and Bryan Perozzi. 2024.
\newblock \href {https://openreview.net/forum?id=IuXR1CCrSi} {Talk like a graph: Encoding graphs for large language models}.
\newblock In \emph{The Twelfth International Conference on Learning Representations}.

\bibitem[{Fey and Lenssen(2019)}]{Fey/Lenssen/2019}
Matthias Fey and Jan~E. Lenssen. 2019.
\newblock Fast graph representation learning with {PyTorch Geometric}.
\newblock In \emph{ICLR Workshop on Representation Learning on Graphs and Manifolds}.

\bibitem[{Gori et~al.(2005)Gori, Monfardini, and Scarselli}]{Gori:ea:2005}
M.~Gori, G.~Monfardini, and F.~Scarselli. 2005.
\newblock \href {https://doi.org/10.1109/IJCNN.2005.1555942} {A new model for learning in graph domains}.
\newblock In \emph{Proceedings. 2005 IEEE International Joint Conference on Neural Networks, 2005.}, volume~2, pages 729--734 vol. 2.

\bibitem[{Guu et~al.(2020)Guu, Lee, Tung, Pasupat, and Chang}]{guu2020retrieval}
Kelvin Guu, Kenton Lee, Zora Tung, Panupong Pasupat, and Mingwei Chang. 2020.
\newblock Retrieval augmented language model pre-training.
\newblock In \emph{International conference on machine learning}, pages 3929--3938. PMLR.

\bibitem[{Hu et~al.(2022)Hu, yelong shen, Wallis, Allen-Zhu, Li, Wang, Wang, and Chen}]{hu2022lora}
Edward~J Hu, yelong shen, Phillip Wallis, Zeyuan Allen-Zhu, Yuanzhi Li, Shean Wang, Lu~Wang, and Weizhu Chen. 2022.
\newblock \href {https://openreview.net/forum?id=nZeVKeeFYf9} {Lo{RA}: Low-rank adaptation of large language models}.
\newblock In \emph{International Conference on Learning Representations}.

\bibitem[{Huang et~al.(2024)Huang, Zhang, Mei, and Ma}]{huang2024llms}
Jin Huang, Xingjian Zhang, Qiaozhu Mei, and Jiaqi Ma. 2024.
\newblock \href {https://arxiv.org/abs/2309.16595} {Can llms effectively leverage graph structural information through prompts, and why?}
\newblock \emph{Preprint}, arXiv:2309.16595.

\bibitem[{Iyyer et~al.(2017)Iyyer, Yih, and Chang}]{iyyer-etal-2017-search}
Mohit Iyyer, Wen-tau Yih, and Ming-Wei Chang. 2017.
\newblock \href {https://doi.org/10.18653/v1/P17-1167} {Search-based neural structured learning for sequential question answering}.
\newblock In \emph{Proceedings of the 55th Annual Meeting of the Association for Computational Linguistics (Volume 1: Long Papers)}, pages 1821--1831, Vancouver, Canada. Association for Computational Linguistics.

\bibitem[{Izacard et~al.(2024)Izacard, Lewis, Lomeli, Hosseini, Petroni, Schick, Dwivedi-Yu, Joulin, Riedel, and Grave}]{10.5555/3648699.3648950}
Gautier Izacard, Patrick Lewis, Maria Lomeli, Lucas Hosseini, Fabio Petroni, Timo Schick, Jane Dwivedi-Yu, Armand Joulin, Sebastian Riedel, and Edouard Grave. 2024.
\newblock Atlas: few-shot learning with retrieval augmented language models.
\newblock \emph{J. Mach. Learn. Res.}, 24(1).

\bibitem[{Jacobs et~al.(1991)Jacobs, Jordan, Nowlan, and Hinton}]{jacobs1991adaptive}
Robert~A Jacobs, Michael~I Jordan, Steven~J Nowlan, and Geoffrey~E Hinton. 1991.
\newblock Adaptive mixtures of local experts.
\newblock \emph{Neural computation}, 3(1):79--87.

\bibitem[{Jain and Lapata(2021)}]{10.1162/tacl_a_00422}
Parag Jain and Mirella Lapata. 2021.
\newblock \href {https://doi.org/10.1162/tacl_a_00422} {{Memory-Based Semantic Parsing}}.
\newblock \emph{Transactions of the Association for Computational Linguistics}, 9:1197--1212.

\bibitem[{Jain and Lapata(2023)}]{jain-lapata-2023-conversational}
Parag Jain and Mirella Lapata. 2023.
\newblock \href {https://doi.org/10.18653/v1/2023.emnlp-main.535} {Conversational semantic parsing using dynamic context graphs}.
\newblock In \emph{Proceedings of the 2023 Conference on Empirical Methods in Natural Language Processing}, pages 8667--8679, Singapore. Association for Computational Linguistics.

\bibitem[{Jiang et~al.(2023)Jiang, Sablayrolles, Mensch, Bamford, Chaplot, Casas, Bressand, Lengyel, Lample, Saulnier et~al.}]{jiang2023mistral}
Albert~Q Jiang, Alexandre Sablayrolles, Arthur Mensch, Chris Bamford, Devendra~Singh Chaplot, Diego de~las Casas, Florian Bressand, Gianna Lengyel, Guillaume Lample, Lucile Saulnier, et~al. 2023.
\newblock Mistral 7b.
\newblock \emph{arXiv preprint arXiv:2310.06825}.

\bibitem[{Kacupaj et~al.(2021)Kacupaj, Plepi, Singh, Thakkar, Lehmann, and Maleshkova}]{kacupaj-etal-2021-conversational}
Endri Kacupaj, Joan Plepi, Kuldeep Singh, Harsh Thakkar, Jens Lehmann, and Maria Maleshkova. 2021.
\newblock \href {https://doi.org/10.18653/v1/2021.eacl-main.72} {Conversational question answering over knowledge graphs with transformer and graph attention networks}.
\newblock In \emph{Proceedings of the 16th Conference of the European Chapter of the Association for Computational Linguistics: Main Volume}, pages 850--862, Online. Association for Computational Linguistics.

\bibitem[{Khandelwal et~al.(2020)Khandelwal, Levy, Jurafsky, Zettlemoyer, and Lewis}]{Khandelwal2020Generalization}
Urvashi Khandelwal, Omer Levy, Dan Jurafsky, Luke Zettlemoyer, and Mike Lewis. 2020.
\newblock \href {https://openreview.net/forum?id=HklBjCEKvH} {Generalization through memorization: Nearest neighbor language models}.
\newblock In \emph{International Conference on Learning Representations}.

\bibitem[{Kingma and Ba(2015)}]{adam}
Diederik~P. Kingma and Jimmy Ba. 2015.
\newblock \href {http://arxiv.org/abs/1412.6980} {Adam: {A} method for stochastic optimization}.
\newblock In \emph{3rd International Conference on Learning Representations, {ICLR} 2015, San Diego, CA, USA, May 7-9, 2015, Conference Track Proceedings}.

\bibitem[{Levenshtein(1965)}]{Levenshtein1965BinaryCC}
Vladimir~I. Levenshtein. 1965.
\newblock Binary codes capable of correcting deletions, insertions, and reversals.
\newblock \emph{Soviet physics. Doklady}, 10:707--710.

\bibitem[{Li et~al.(2023)Li, Li, Savarese, and Hoi}]{li2023blip}
Junnan Li, Dongxu Li, Silvio Savarese, and Steven Hoi. 2023.
\newblock Blip-2: Bootstrapping language-image pre-training with frozen image encoders and large language models.
\newblock In \emph{International conference on machine learning}, pages 19730--19742. PMLR.

\bibitem[{Liu et~al.(2023)Liu, Li, Wu, and Lee}]{NEURIPS2023_6dcf277e}
Haotian Liu, Chunyuan Li, Qingyang Wu, and Yong~Jae Lee. 2023.
\newblock \href {https://proceedings.neurips.cc/paper_files/paper/2023/file/6dcf277ea32ce3288914faf369fe6de0-Paper-Conference.pdf} {Visual instruction tuning}.
\newblock In \emph{Advances in Neural Information Processing Systems}, volume~36, pages 34892--34916. Curran Associates, Inc.

\bibitem[{McCoy et~al.(2023)McCoy, Yao, Friedman, Hardy, and Griffiths}]{mccoy2023embers}
R.~Thomas McCoy, Shunyu Yao, Dan Friedman, Matthew Hardy, and Thomas~L. Griffiths. 2023.
\newblock \href {https://arxiv.org/abs/2309.13638} {Embers of autoregression: Understanding large language models through the problem they are trained to solve}.
\newblock \emph{Preprint}, arXiv:2309.13638.

\bibitem[{Mueller et~al.(2019)Mueller, Piccinno, Shaw, Nicosia, and Altun}]{mueller-etal-2019-answering}
Thomas Mueller, Francesco Piccinno, Peter Shaw, Massimo Nicosia, and Yasemin Altun. 2019.
\newblock \href {https://doi.org/10.18653/v1/D19-1603} {Answering conversational questions on structured data without logical forms}.
\newblock In \emph{Proceedings of the 2019 Conference on Empirical Methods in Natural Language Processing and the 9th International Joint Conference on Natural Language Processing (EMNLP-IJCNLP)}, pages 5902--5910, Hong Kong, China. Association for Computational Linguistics.

\bibitem[{Parikh et~al.(2020)Parikh, Wang, Gehrmann, Faruqui, Dhingra, Yang, and Das}]{parikh-etal-2020-totto}
Ankur Parikh, Xuezhi Wang, Sebastian Gehrmann, Manaal Faruqui, Bhuwan Dhingra, Diyi Yang, and Dipanjan Das. 2020.
\newblock \href {https://doi.org/10.18653/v1/2020.emnlp-main.89} {{ToTTo}: A controlled table-to-text generation dataset}.
\newblock In \emph{Proceedings of the 2020 Conference on Empirical Methods in Natural Language Processing (EMNLP)}, pages 1173--1186, Online. Association for Computational Linguistics.

\bibitem[{Paszke et~al.(2019)Paszke, Gross, Massa, Lerer, Bradbury, Chanan, Killeen, Lin, Gimelshein, Antiga et~al.}]{paszke2019pytorch}
Adam Paszke, Sam Gross, Francisco Massa, Adam Lerer, James Bradbury, Gregory Chanan, Trevor Killeen, Zeming Lin, Natalia Gimelshein, Luca Antiga, et~al. 2019.
\newblock Pytorch: An imperative style, high-performance deep learning library.
\newblock \emph{Advances in neural information processing systems}, 32.

\bibitem[{Perez-Beltrachini et~al.(2023)Perez-Beltrachini, Jain, Monti, and Lapata}]{perez-beltrachini-etal-2023-semantic}
Laura Perez-Beltrachini, Parag Jain, Emilio Monti, and Mirella Lapata. 2023.
\newblock \href {https://doi.org/10.18653/v1/2023.eacl-main.184} {Semantic parsing for conversational question answering over knowledge graphs}.
\newblock In \emph{Proceedings of the 17th Conference of the European Chapter of the Association for Computational Linguistics}, pages 2507--2522, Dubrovnik, Croatia. Association for Computational Linguistics.

\bibitem[{Perozzi et~al.(2024)Perozzi, Fatemi, Zelle, Tsitsulin, Kazemi, Al-Rfou, and Halcrow}]{perozzi2024let}
Bryan Perozzi, Bahare Fatemi, Dustin Zelle, Anton Tsitsulin, Mehran Kazemi, Rami Al-Rfou, and Jonathan Halcrow. 2024.
\newblock Let your graph do the talking: Encoding structured data for llms.
\newblock \emph{arXiv preprint arXiv:2402.05862}.

\bibitem[{Raffel et~al.(2020)Raffel, Shazeer, Roberts, Lee, Narang, Matena, Zhou, Li, and Liu}]{JMLR:v21:20-074}
Colin Raffel, Noam Shazeer, Adam Roberts, Katherine Lee, Sharan Narang, Michael Matena, Yanqi Zhou, Wei Li, and Peter~J. Liu. 2020.
\newblock \href {http://jmlr.org/papers/v21/20-074.html} {Exploring the limits of transfer learning with a unified text-to-text transformer}.
\newblock \emph{Journal of Machine Learning Research}, 21(140):1--67.

\bibitem[{Rajpurkar et~al.(2016)Rajpurkar, Zhang, Lopyrev, and Liang}]{rajpurkar-etal-2016-squad}
Pranav Rajpurkar, Jian Zhang, Konstantin Lopyrev, and Percy Liang. 2016.
\newblock \href {https://doi.org/10.18653/v1/D16-1264} {{SQ}u{AD}: 100,000+ questions for machine comprehension of text}.
\newblock In \emph{Proceedings of the 2016 Conference on Empirical Methods in Natural Language Processing}, pages 2383--2392, Austin, Texas. Association for Computational Linguistics.

\bibitem[{Raposo et~al.(2022)Raposo, Ribeiro, Martins, and Coheur}]{raposo2022question}
Gon{\c{c}}alo Raposo, Rui Ribeiro, Bruno Martins, and Lu{\'\i}sa Coheur. 2022.
\newblock Question rewriting? assessing its importance for conversational question answering.
\newblock In \emph{European Conference on Information Retrieval}, pages 199--206. Springer.

\bibitem[{Reddy et~al.(2019)Reddy, Chen, and Manning}]{reddy-etal-2019-coqa}
Siva Reddy, Danqi Chen, and Christopher~D. Manning. 2019.
\newblock \href {https://doi.org/10.1162/tacl_a_00266} {{C}o{QA}: A conversational question answering challenge}.
\newblock \emph{Transactions of the Association for Computational Linguistics}, 7:249--266.

\bibitem[{Reimers and Gurevych(2019)}]{reimers-gurevych-2019-sentence}
Nils Reimers and Iryna Gurevych. 2019.
\newblock \href {https://doi.org/10.18653/v1/D19-1410} {Sentence-{BERT}: Sentence embeddings using {S}iamese {BERT}-networks}.
\newblock In \emph{Proceedings of the 2019 Conference on Empirical Methods in Natural Language Processing and the 9th International Joint Conference on Natural Language Processing (EMNLP-IJCNLP)}, pages 3982--3992, Hong Kong, China. Association for Computational Linguistics.

\bibitem[{Robertson and Zaragoza(2009)}]{bm25}
Stephen Robertson and Hugo Zaragoza. 2009.
\newblock \href {https://doi.org/10.1561/1500000019} {The probabilistic relevance framework: Bm25 and beyond}.
\newblock \emph{Found. Trends Inf. Retr.}, 3(4):333–389.

\bibitem[{Saha et~al.(2018)Saha, Pahuja, Khapra, Sankaranarayanan, and Chandar}]{10.5555/3504035.3504122}
Amrita Saha, Vardaan Pahuja, Mitesh~M. Khapra, Karthik Sankaranarayanan, and Sarath Chandar. 2018.
\newblock Complex sequential question answering: towards learning to converse over linked question answer pairs with a knowledge graph.
\newblock In \emph{Proceedings of the Thirty-Second AAAI Conference on Artificial Intelligence and Thirtieth Innovative Applications of Artificial Intelligence Conference and Eighth AAAI Symposium on Educational Advances in Artificial Intelligence}, AAAI'18/IAAI'18/EAAI'18. AAAI Press.

\bibitem[{Scarselli et~al.(2009)Scarselli, Gori, Tsoi, Hagenbuchner, and Monfardini}]{Scarselli:ea:2009}
Franco Scarselli, Marco Gori, Ah~Chung Tsoi, Markus Hagenbuchner, and Gabriele Monfardini. 2009.
\newblock \href {https://doi.org/10.1109/TNN.2008.2005605} {The graph neural network model}.
\newblock \emph{IEEE Transactions on Neural Networks}, 20(1):61--80.

\bibitem[{Shazeer et~al.(2017)Shazeer, Mirhoseini, Maziarz, Davis, Le, Hinton, and Dean}]{shazeer2017outrageously}
Noam Shazeer, Azalia Mirhoseini, Krzysztof Maziarz, Andy Davis, Quoc Le, Geoffrey Hinton, and Jeff Dean. 2017.
\newblock Outrageously large neural networks: The sparsely-gated mixture-of-experts layer.
\newblock \emph{arXiv preprint arXiv:1701.06538}.

\bibitem[{Shen et~al.(2019)Shen, Geng, Qin, Guo, Tang, Duan, Long, and Jiang}]{shen-etal-2019-multi}
Tao Shen, Xiubo Geng, Tao Qin, Daya Guo, Duyu Tang, Nan Duan, Guodong Long, and Daxin Jiang. 2019.
\newblock \href {https://doi.org/10.18653/v1/D19-1248} {Multi-task learning for conversational question answering over a large-scale knowledge base}.
\newblock In \emph{Proceedings of the 2019 Conference on Empirical Methods in Natural Language Processing and the 9th International Joint Conference on Natural Language Processing (EMNLP-IJCNLP)}, pages 2442--2451, Hong Kong, China. Association for Computational Linguistics.

\bibitem[{Singh and Strouse(2024)}]{singh2024tokenization}
Aaditya~K. Singh and DJ~Strouse. 2024.
\newblock \href {https://arxiv.org/abs/2402.14903} {Tokenization counts: the impact of tokenization on arithmetic in frontier llms}.
\newblock \emph{Preprint}, arXiv:2402.14903.

\bibitem[{Sun et~al.(2023)Sun, Qi, Zhang, Liu, Wang, and Huang}]{sun-etal-2023-tokenization}
Kaiser Sun, Peng Qi, Yuhao Zhang, Lan Liu, William Wang, and Zhiheng Huang. 2023.
\newblock \href {https://doi.org/10.18653/v1/2023.findings-emnlp.887} {Tokenization consistency matters for generative models on extractive {NLP} tasks}.
\newblock In \emph{Findings of the Association for Computational Linguistics: EMNLP 2023}, pages 13300--13310, Singapore. Association for Computational Linguistics.

\bibitem[{Sun and Chai(2007)}]{sun2007discourse}
Mingyu Sun and Joyce~Y Chai. 2007.
\newblock Discourse processing for context question answering based on linguistic knowledge.
\newblock \emph{Knowledge-Based Systems}, 20(6):511--526.

\bibitem[{Touvron et~al.(2023)Touvron, Martin, Stone, Albert, Almahairi, Babaei, Bashlykov, Batra, Bhargava, Bhosale et~al.}]{touvron2023llama}
Hugo Touvron, Louis Martin, Kevin Stone, Peter Albert, Amjad Almahairi, Yasmine Babaei, Nikolay Bashlykov, Soumya Batra, Prajjwal Bhargava, Shruti Bhosale, et~al. 2023.
\newblock Llama 2: Open foundation and fine-tuned chat models.
\newblock \emph{arXiv preprint arXiv:2307.09288}.

\bibitem[{Velickovic et~al.(2018)Velickovic, Cucurull, Casanova, Romero, Li{\`{o}}, and Bengio}]{gat2018graph}
Petar Velickovic, Guillem Cucurull, Arantxa Casanova, Adriana Romero, Pietro Li{\`{o}}, and Yoshua Bengio. 2018.
\newblock \href {https://openreview.net/forum?id=rJXMpikCZ} {Graph attention networks}.
\newblock In \emph{6th International Conference on Learning Representations, {ICLR} 2018, Vancouver, BC, Canada, April 30 - May 3, 2018, Conference Track Proceedings}. OpenReview.net.

\bibitem[{von Werra et~al.(2020)von Werra, Belkada, Tunstall, Beeching, Thrush, Lambert, and Huang}]{vonwerra2022trl}
Leandro von Werra, Younes Belkada, Lewis Tunstall, Edward Beeching, Tristan Thrush, Nathan Lambert, and Shengyi Huang. 2020.
\newblock Trl: Transformer reinforcement learning.
\newblock \url{https://github.com/huggingface/trl}.

\bibitem[{Voskarides et~al.(2020)Voskarides, Li, Ren, Kanoulas, and de~Rijke}]{voskarides2020query}
Nikos Voskarides, Dan Li, Pengjie Ren, Evangelos Kanoulas, and Maarten de~Rijke. 2020.
\newblock Query resolution for conversational search with limited supervision.
\newblock In \emph{Proceedings of the 43rd International ACM SIGIR conference on research and development in Information Retrieval}, pages 921--930.

\bibitem[{Vrande\v{c}i\'{c} and Kr\"{o}tzsch(2014)}]{10.1145/2629489}
Denny Vrande\v{c}i\'{c} and Markus Kr\"{o}tzsch. 2014.
\newblock \href {https://doi.org/10.1145/2629489} {Wikidata: a free collaborative knowledgebase}.
\newblock \emph{Commun. ACM}, 57(10):78–85.

\bibitem[{Wang et~al.(2024)Wang, Feng, He, Tan, Han, and Tsvetkov}]{wang2024can}
Heng Wang, Shangbin Feng, Tianxing He, Zhaoxuan Tan, Xiaochuang Han, and Yulia Tsvetkov. 2024.
\newblock Can language models solve graph problems in natural language?
\newblock \emph{Advances in Neural Information Processing Systems}, 36.

\bibitem[{Wang et~al.(2023)Wang, Kordi, Mishra, Liu, Smith, Khashabi, and Hajishirzi}]{wang-etal-2023-self-instruct}
Yizhong Wang, Yeganeh Kordi, Swaroop Mishra, Alisa Liu, Noah~A. Smith, Daniel Khashabi, and Hannaneh Hajishirzi. 2023.
\newblock \href {https://doi.org/10.18653/v1/2023.acl-long.754} {Self-instruct: Aligning language models with self-generated instructions}.
\newblock In \emph{Proceedings of the 61st Annual Meeting of the Association for Computational Linguistics (Volume 1: Long Papers)}, pages 13484--13508, Toronto, Canada. Association for Computational Linguistics.

\bibitem[{Wolf et~al.(2020)Wolf, Debut, Sanh, Chaumond, Delangue, Moi, Cistac, Rault, Louf, Funtowicz, Davison, Shleifer, von Platen, Ma, Jernite, Plu, Xu, Le~Scao, Gugger, Drame, Lhoest, and Rush}]{wolf-etal-2020-transformers}
Thomas Wolf, Lysandre Debut, Victor Sanh, Julien Chaumond, Clement Delangue, Anthony Moi, Pierric Cistac, Tim Rault, Remi Louf, Morgan Funtowicz, Joe Davison, Sam Shleifer, Patrick von Platen, Clara Ma, Yacine Jernite, Julien Plu, Canwen Xu, Teven Le~Scao, Sylvain Gugger, Mariama Drame, Quentin Lhoest, and Alexander Rush. 2020.
\newblock \href {https://doi.org/10.18653/v1/2020.emnlp-demos.6} {Transformers: State-of-the-art natural language processing}.
\newblock In \emph{Proceedings of the 2020 Conference on Empirical Methods in Natural Language Processing: System Demonstrations}, pages 38--45, Online. Association for Computational Linguistics.

\bibitem[{Ye et~al.(2023)Ye, Zhang, Wang, Xu, and Zhang}]{ye2023natural}
Ruosong Ye, Caiqi Zhang, Runhui Wang, Shuyuan Xu, and Yongfeng Zhang. 2023.
\newblock Natural language is all a graph needs.
\newblock \emph{arXiv preprint arXiv:2308.07134}.

\end{thebibliography}
\appendix
\section{Prompt Description}
\label{app:sec:prompts}

Figure~\ref{fig:zero_shot} shows an example prompt for the Mistral-7B model without graph embeddings (see Mistral-7B zero-shot  in Table~\ref{tab:mistral_results}). The prompt includes a sequence of retrieved and ranked pieces of evidence, each encapsulated within \texttt{<evidence>--</evidence>} tags. We represent the past interaction $\text{I}[:t - 1]$ as a series of question and answer pairs. The same prompt is used for fine-tuning (see Mistral-7B + FT in Table~\ref{tab:mistral_results}) with  the subsequent response as the gold output tokens (see  Section~\ref{sec:training} for details).

Figure~\ref{fig:graph_prompt} shows an example prompt for the graph-based model (all model variants with +Graph in Table~\ref{tab:mistral_results}). The prompt consists of three parts, the initial instructions which we refer to as~$\mathsf{P}_{\text{prefix}}$, a sequence of graph node embeddings represented as \texttt{graph\_node\_embedding}, and the conversational query which we denote as~$\mathsf{P}_{\text{suffix}}$.

\section{Training Details}
\label{app:sec:hyper}
Table~\ref{tab:hyperparams} list the hyper-parameters employed to train our model. Implementation details are discussed in Section~\ref{sec:experimental-setup}. During the fine-tuning of the base language model, only the query, key, and value projection parameters are updated.

\begin{table}[ht]
    \centering
    \begin{tabular}{l|p{4cm}}
    \textbf{Parameter} & \textbf{Value} \\
    \hline
    Graph layers & 2 \\
    Graph heads & 2 \\
    Lora rank & 128 \\
    Lora $\alpha$ & 32 \\
    Lora dropout & 0.05 \\
    GAT Dropout & 0.5 \\
    Optimizer & Adam~\cite{adam} \\
    Learning rate & 5e-5 \\
    Batch size & 1 \\
    Gradient accumulation & 4 \\
    \end{tabular}
    \caption{Hyperparameter values used for our model.}
    \label{tab:hyperparams}
\end{table}

\begin{figure*}
\centering
 \begin{tcolorbox}[colback=white, colframe=blue!30, left=2pt,  coltitle=black, title=\textbf{Prompt}: Mistral-7B zero shot and fine-tuned without graph embeddings, halign=flush left]
    \footnotesize
[INST] \\ \vspace{3mm}
You are a helpful assistant. Using the following facts: \\\vspace{3mm}
<evidence>\textcolor{darkgray}{Kid A, publication, 2 October 2000}</evidence>\\\vspace{3mm}
<evidence>\textcolor{darkgray}{Rolling Stone, Editor, Noah Shachtman}</evidence>\\\vspace{3mm}
<evidence>\textcolor{darkgray}{Rolling Stone, Catgories, Popular culture}</evidence>\\\vspace{3mm}
<evidence>\textcolor{darkgray}{Publication Fact, Country UK, Accolade The 100 Best Albums of the 2000s, Year 2010, Rank 7}</evidence>\\\vspace{3mm}
<evidence>\textcolor{darkgray}{Publication Rolling Stone, Country US, Accolade The 100 Best Albums of the decade, Year 2009, Rank~1}</evidence>\\\vspace{3mm}
<evidence>\textcolor{darkgray}{Rolling Stone was founded in San Francisco in 1967 by Jann Wenner and Ralph J. Gleason.}</evidence>\\\vspace{3mm}
Answer the following conversational query as a simple key fact without description:\\\vspace{3mm}

[/INST]\\\vspace{3mm}
\begin{tabular}{ll}
Question: & What is the release date of album Kid A?\\
Answer: & 2 October 2000\\
Question: &Fact Rank?\\
Answer: &7\\
Question: &Ranking on Rolling Stone in 2009?\\
Answer: & \\
\end{tabular}

    \end{tcolorbox}
      \caption{Example prompt for models which do not employ graph embeddings. Only a few relevant pieces of evidence are shown, for the sake of brevity.}
    \label{fig:zero_shot}
\end{figure*}

\begin{figure*}
\centering
 \begin{tcolorbox}[colback=white, colframe=blue!30, left=2pt,  coltitle=black, title=\textbf{Prompt}: Mistral-7B fine-tuned with graph embeddings, halign=flush left]
    \footnotesize
   
    [INST] \\ \vspace{3mm}You are a helpful assistant. Using the following facts: \\ \vspace{3mm}
\textcolor{darkgray}{[\texttt{graph\_node\_embedding\_1, graph\_node\_embedding\_2, ... , graph\_node\_embedding\_n]}} \\ \vspace{3mm}
Answer the following conversational query as a simple key fact without description:\\ \vspace{3mm}
[/INST]\\ \vspace{3mm}
\begin{tabular}{ll}
Question: & What is the release date of album Kid A?\\
Answer: & 2 October 2000\\
Question: &Fact Rank?\\
Answer: & 7\\
Question: &Ranking on Rolling Stone in 2009?\\
Answer: &\\
\end{tabular}
\end{tcolorbox}
    \caption{Example prompt for graph-based models. We use~$\mathsf{P}_{\text{prefix}}$ and~$\mathsf{P}_{\text{suffix}}$ to denote the instruction before and after the  \textcolor{darkgray}{\texttt{graph\_node\_embeddings}} respectively. The umber of graph node embeddings is dynamic and varies based on evidence that has been retrieved.}
    \label{fig:graph_prompt}
\end{figure*}

\end{document}